\documentclass[10pt,twocolumn,letterpaper]{article}

 \usepackage{cvpr}              % To produce the CAMERA-READY version
\usepackage[T1]{fontenc}

\definecolor{cvprblue}{rgb}{0.21,0.49,0.74}
\usepackage[pagebackref,breaklinks,colorlinks,allcolors=cvprblue]{hyperref}
\usepackage{tcolorbox}
\usepackage{graphicx}
\usepackage{multirow} 
\usepackage{cuted}
\usepackage{capt-of} 
\def\confName{CVPR}
\def\confYear{2027}

\title{SpatialScout-V2: Can VLM Agents Actively Explore Real-World \\ 3D Scenes and Localize Objects?}

\author{
Zhaoxi Zhang$^{1}$\thanks{Project Lead.} \thanks{Equal Contribution}  \quad
Hao Liang$^{1}$\footnotemark[2] \quad
Linxuan Wu$^{2}$\footnotemark[2] \quad
Yiming Xu$^{1}$\footnotemark[2] \quad
Yifu Yuan$^{2}$\footnotemark[2] \quad
Kun Liang$^{1}$ \\
Chenming Tang$^{1}$ \quad
Meiyi Qiang$^{1}$ \quad
Wentao Zhang$^{1}$ \quad
Weikang Li$^{1}$\thanks{Corresponding author.} \quad
Yunfang Wu$^{1}$\footnotemark[3] \\
$^{1}$\,Peking University \qquad
$^{2}$\,Tianjin University \\
\href{mailto:zhaoxizhang25@stu.pku.edu.cn}{zhaoxizhang25@stu.pku.edu.cn}
}

\begin{document}
\maketitle

% \begin{figure*}[t]
%     \centering
%     \includegraphics[width=1\linewidth]{sec//figs/viewscout_overview.pdf}
%     \caption{Overview of ViewScout, a benchmark consisting of diverse indoor scenes and objects.}
%     \label{fig:placeholder}
% \end{figure*}

\begin{strip}
    \centering
    \vspace{-65pt}
    \includegraphics[width=\textwidth]{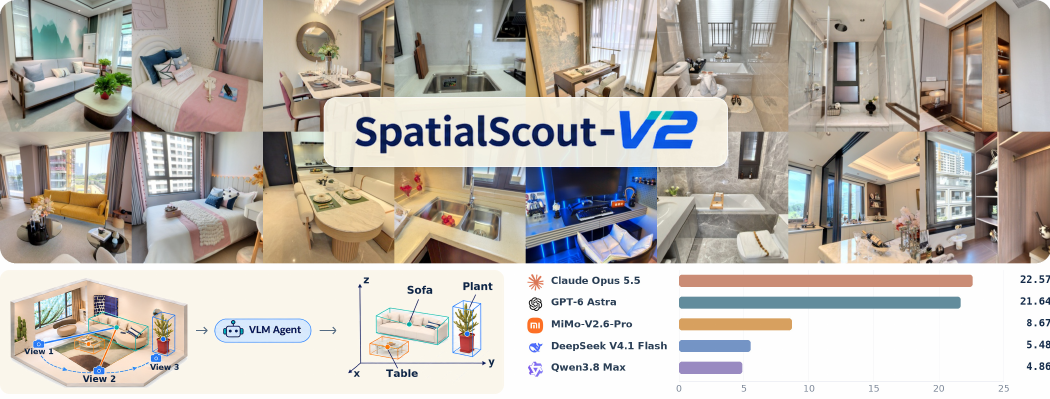}
    \captionof{figure}{Overview of SpatialScout-V2, a benchmark built from real-world scenes for 3D object localization. A VLM agent controls the camera, conducts automated reasoning, and submits the world coordinates of all spotted objects.}
    \label{fig:spatialscout-teaser}
\end{strip}

\begin{abstract}
Spatial intelligence requires agents to acquire visual evidence and integrate partial observations into coherent three-dimensional representations.
We introduce \textbf{SpatialScout-V2}, which extends SpatialScout's active mapping task from simulation to real residential scenes through three advances: real visual observations, preservation of original scene content, and interactive exploration of recorded views.
Built from 100 homes in Realsee3D, the benchmark covers 3,171 target objects across 29 categories.
Given a fixed target inventory, agents autonomously select recorded viewpoints and viewing directions, integrate calibrated RGB observations, and iteratively estimate and revise object centers within a fixed interaction budget.
We evaluate the localization accuracy and completeness of object-center maps using mean IoU, IoU-threshold hit rates, and target coverage.
Evaluating seven frontier vision-language models (VLMs), we find that six achieve an all-target mean IoU below 20\%.
Even the best-performing model, Claude Opus~5.5, reaches only 22.52\%, with just 18.36\% of targets exceeding an IoU of 0.5.
Further analysis shows that broad submission coverage can coexist with limited localization accuracy, while target omissions further reduce overall scores.
These results highlight substantial challenges in constructing accurate and complete object-center maps through active exploration of real scenes.
\end{abstract}

\section{Introduction}
\label{sec:introduction}

Spatial intelligence enables agents to understand the spatial layout of their surroundings and use that understanding to guide their actions. When observations are partial, this understanding must be built incrementally~\cite{theoryofspace}. Agents need to decide which regions warrant further exploration, recognize the same objects across viewpoints, and integrate local position estimates into a shared three-dimensional representation~\cite{conceptgraphs}. As evidence accumulates, this representation must also be revised. Spatial intelligence thus couples exploration and representation: the current scene understanding guides subsequent observations, which in turn update that understanding~\cite{bajcsy1988active,naruto}.

Evaluating this capability requires examining both how agents acquire evidence and how they use it to construct spatial representations. Spatial question answering can test specific judgments about distance, direction, and object relations~\cite{vsi}, but correctly answering individual questions does not establish whether vision-language model (VLM) agents can complete long-horizon spatial interaction tasks. SpatialScout~\cite{spatialscout} introduced a concrete evaluation of this capability in simulation: agents autonomously explore indoor scenes and reconstruct their spatial layouts by estimating three-dimensional object positions. This task provides a foundation for evaluating active spatial construction, but agents' performance on this task with real-world observations remains to be examined. We therefore ask: can agents construct accurate and complete object-center maps from observations of real homes?

SpatialScout-V2 extends SpatialScout's active spatial construction task from simulation to real residential captures. While retaining autonomous observation, evidence integration, and map revision, v2 advances the evaluation setting in three respects:

\begin{enumerate}
\item \textbf{Real visual observations.} We replace images rendered from simulated assets with views from real captures, preserving the appearance of objects and the lighting conditions of actual homes.
\item \textbf{Preservation of original scene content.} Whereas v1 hides unselected candidate objects during rendering~\cite{spatialscout}, v2 retains the original scene content in the captures, including both target and non-target objects and their occlusion relationships. Agents therefore explore and localize objects within their natural scene context.
\item \textbf{Interactive exploration of recorded views.} We organize panoramic captures taken at multiple positions in real homes into an interactive environment. Guided by prior observations and the current map, agents select subsequent viewing positions from the recorded locations and adjust viewing directions to gather further evidence and refine spatial estimates.
\end{enumerate}

These advances combine real visual observations, autonomous evidence acquisition, and explicit spatial evaluation to assess whether agents can construct accurate and complete spatial representations. VSI-Bench~\cite{vsi} evaluates spatial reasoning over fixed video observations, while OpenEQA's active setting~\cite{openeqa} and E3VS-Bench~\cite{e3vs} allow agents to acquire evidence for question answering. SpatialScout-V2 instead evaluates the object-center map progressively constructed through interaction, jointly measuring localization accuracy and target coverage. Agents must select informative viewpoints, associate evidence across views, and continually refine position estimates amid non-target objects and occlusions, balancing map completeness and accuracy within a limited interaction budget.

SpatialScout-V2 uses 100 real residential scenes from Realsee3D~\cite{Li2025realsee3d_data,li2026argus}, covering 3,171 target objects across 29 categories. The evaluation environment combines registered panoramic observations with reference geometry derived from real captures, enabling quantitative evaluation of SpatialScout's active mapping task in real scenes. Agents have access to a fixed target inventory, available camera positions, and calibrated RGB views. Each episode allows up to 120 model turns, during which agents select recorded viewpoints, adjust viewing directions, acquire images, and submit or revise three-dimensional object centers. Agents determine their own observation sequences, so each new view can inform both map updates and subsequent exploration. A triangulation tool supports geometric estimation from agent-selected point correspondences, while depth maps and reference annotations remain unavailable to the agents. We use the source dataset's multi-view 3D annotations as evaluation references and assess localization relative to object size using 3D bounding-box intersection over union (IoU). For scoring, each submitted identifier's reference-sized box is translated to the predicted center and matched one-to-one within its category. Scores over both all targets and submitted targets distinguish the effect of target omissions from the localization qualitya.

Our evaluation of seven VLM agents reveals substantial limitations in spatial localization under this real-scene protocol. Among completed episodes that satisfy the scoring criteria, Claude Opus~5.5 achieves the highest all-target mean IoU at 22.52\%; only 18.36\% of targets exceed an IoU of 0.5, and just 0.19\% exceed 0.9. Its submitted-target mean IoU is nearly identical at 22.58\%, showing that broad submission coverage can coexist with limited localization accuracy. GPT-6 Astra reaches a mean IoU of 21.65\% over submitted targets but 15.50\% over all targets, reflecting the additional effect of target omissions. Tool-use analysis further reveals differences in how agents allocate interactions to image acquisition and triangulation, as well as repeated operations. These findings underscore the need to evaluate agents' acquisition and integration of visual evidence together with the accuracy and completeness of their final spatial estimates.

Our contributions are threefold. First, we extend SpatialScout from active mapping in simulation to interactive mapping from real residential captures, retaining autonomous exploration and map revision while incorporating real appearance, lighting, and non-target scene content into the evaluation. Second, we combine autonomous view selection from registered panoramas with source-dataset 3D annotations to quantify localization accuracy and inventory coverage across 100 homes and 3,171 target objects. Third, we evaluate seven VLM agents and analyze their localization results, coverage, and tool-use behavior, revealing persistent difficulty in constructing precise object-center maps even when submission coverage is high.

\begin{figure*}[t]
    \centering
    \includegraphics[width=1\linewidth]{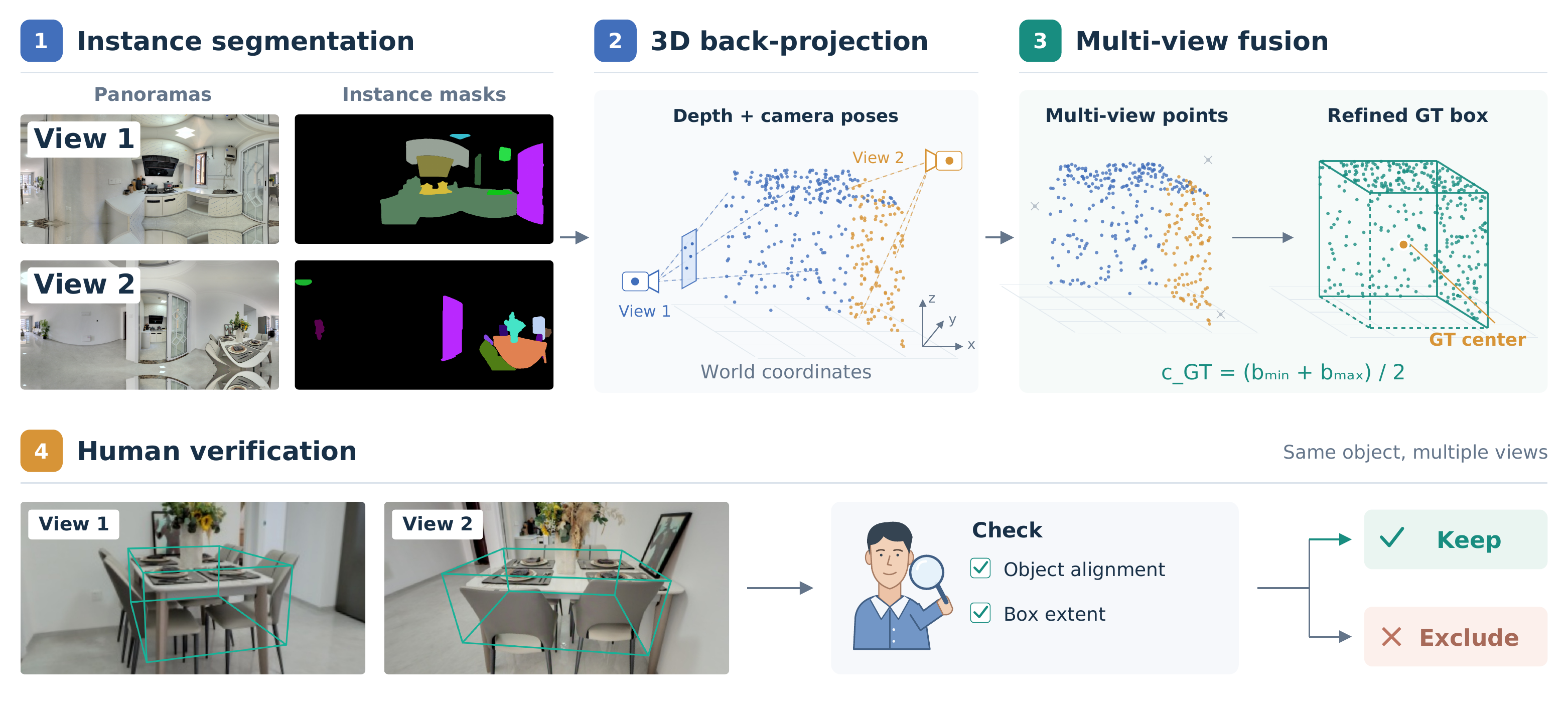}
    \caption{Reference 3D object annotation from real captures. Instance masks are extracted from panoramas, lifted into a common metric frame using depth and camera poses, and fused across views. Geometric refinement produces a scene-level box and its center for each object. Human annotators verify the boxes from at least two viewpoints.}
    \label{fig:construction}
\end{figure*}

\section{Benchmark Construction}
\label{sec:method}
\label{sec:benchmark_construction}

SpatialScout-V2 evaluates active object-center mapping on the 100 test scenes of Realsee3D's real-world subset~\cite{Li2025realsee3d_data,li2026argus}.
Given a target inventory and RGB observations with known camera poses and intrinsics, an agent selects among recorded camera positions and estimates the targets' three-dimensional centers.
The benchmark measures both localization accuracy and coverage of the inventory.

\begin{figure}[t]
    \centering
    \includegraphics[width=1\linewidth]{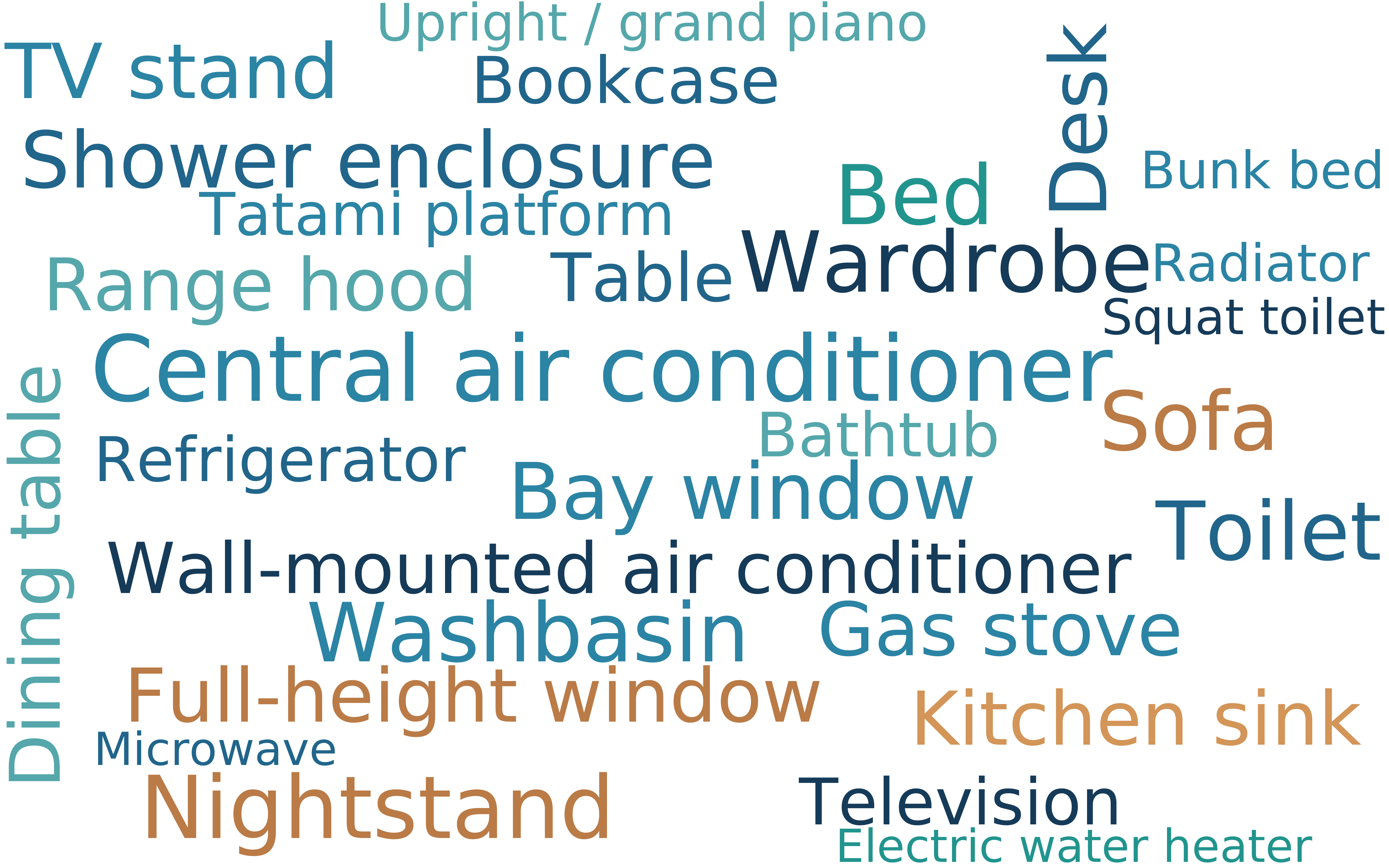}
    \caption{Wordcloud of SpatialScout-V2. The dataset contains 29 types and 3,171 objects from 100 scenes.}
    \label{fig:wordcloud}
\end{figure}

\subsection{Real-World Scene Data}
\label{sec:real_scene_data}

The real-world subset of Realsee3D contains 1,000 residential scenes, partitioned at scene level into training and test sets in a 9:1 ratio~\cite{li2026argus}.
We use its 100 test scenes as a fixed evaluation catalog.
Realsee captures these interiors using tripod-mounted Galois 3D LiDAR cameras.
The released data provide equirectangular RGB panoramas, sparse depth maps, and camera-to-world transformations for each scan position.
Metric depth is recovered using the supplied depth scale. The aligned RGB images and depth maps support the multi-view 3D object annotations described below.

\subsection{Object Annotations}
\label{sec:object_annotations}

The reference boxes are produced by an automatic multi-view annotation pipeline, as depicted in Fig.~\ref{fig:construction}. We manually inspect each box from at least two viewpoints and remove inaccurate annotations. The resulting inventory contains 3,171 objects across 29 categories, as presented in Fig.~\ref{fig:wordcloud}. We use the resulting boxes as evaluation references after visual inspection.

\paragraph{Panoramic instance segmentation.}
A proprietary panoramic segmentation model developed by Realsee \cite{Li2025realsee3d_data} predicts semantic regions, instance masks, and confidence scores directly on the full panoramas.
Low-confidence detections are removed, and masks are refined using geometric and semantic cues.
Mirrors and glass can cause erroneous laser range measurements~\cite{koch2017reflections}.
The annotation pipeline suppresses pixels on these surfaces.
Masks crossing the panorama seam and masks that combine distinct objects are repartitioned before lifting to 3D.

\paragraph{Lifting and association across views.}
Each refined mask is back-projected using the aligned metric depth and the corresponding camera-to-world pose.
Bidirectional reprojection consistency associates observations of the same physical object across views.
Their 3D points are fused, and outlier removal reduces depth noise and points included through segmentation errors.
A gravity-aligned bounding box is fitted to each fused instance.
Fragments with strongly overlapping footprints are merged, and box centers and orientations are refined using the dominant scene axes.
Category-specific size priors, duplicate removal, and correction of category confusions complete the annotation.

The source 3D annotation vocabulary contains 102 indoor object categories drawn from the segmentation model's 130 categories.
Each annotated instance includes its category, 3D box, confidence aggregated over its 2D observations, visible-viewpoint identifiers, and floor index.
These scene-level instances provide a reference for mapping the same object across multiple photographs.

\subsection{Mapping Targets and Coordinates}
\label{sec:mapping_task}

For the $N$ selected target objects, the public inventory is $\mathcal{E}=\{(i,c_i)\}_{i=1}^{N}$, where $i$ is an instance identifier and $c_i$ its category.
Objects with the same category form a matching group.
The agent submits a three-dimensional center $(x,y,z)$ for each target it localizes, and evaluation covers this fixed inventory.

The real-data loader reads the two exported endpoints $\mathbf{a}_i$ and $\mathbf{b}_i$ of each annotated box.
It applies the fixed coordinate conversion $A(x,y,z)^{\top}=(x,z,-y)^{\top}$ to both box endpoints and camera extrinsics.
The scoring bounds and target center are
\begin{equation}
    \begin{aligned}
        \mathbf{b}_i^{-}&=\min(A\mathbf{a}_i,A\mathbf{b}_i),\\
        \mathbf{b}_i^{+}&=\max(A\mathbf{a}_i,A\mathbf{b}_i),\qquad
        \mathbf{p}_i=\frac{\mathbf{b}_i^{-}+\mathbf{b}_i^{+}}{2},
    \end{aligned}
    \label{eq:target_center}
\end{equation}
where minima and maxima are taken componentwise.
These bounds define the world-axis-aligned bounding box (AABB) used for scoring.
For a source camera pose $(R,\mathbf{t})$, the corresponding pose is $(AR,A\mathbf{t})$.
Camera positions, triangulated points, and submitted centers all use this common metric frame, with the dataset origin retained. During inference, the agent selects recorded viewpoints, queries camera poses and intrinsics, and submits estimated object centers.

\section{Evaluation Protocol}
\label{sec:evaluation_protocol}

\subsection{Observations at Recorded Viewpoints}
\label{sec:interaction}

Each model is run on the fixed catalog of 100 scenes with a shared task prompt and tool interface.
An episode starts at the first scan position in lexicographic identifier order, with zero yaw and pitch relative to its panorama.
The agent queries \texttt{list\_camera\_positions} and moves to a recorded location using \texttt{select\_camera\_position}.
All listed positions are selectable by identifier; selection preserves the yaw and pitch relative to the new panorama.
The agent changes the viewing direction with \texttt{rotate\_camera} and obtains an image through \texttt{capture\_image}.
For each scene, the target inventory, available scan positions, initial view, and interaction budget are shared across models.

\paragraph{Perspective observations.}
Each observation is a $512\times512$ perspective projection of the selected panorama, with a $90^{\circ}$ horizontal and vertical field of view.
The projection uses bilinear interpolation and wraps horizontally at the panorama seam.
Yaw is cyclic and pitch is restricted to $[-85^{\circ},85^{\circ}]$.
Frames are stored as PNGs and delivered to the model as JPEG images at the same resolution, with quality 82.
Camera tools return the current pose and pinhole intrinsic matrix.
The camera basis is forward/left/up; image coordinates use zero-indexed pixel centers, with $u$ increasing rightward and $v$ downward.
\begin{table*}[t]
\centering
\small
\setlength{\tabcolsep}{5pt}
\caption{Object-micro localization results on 100 scenes. All values are percentages. We report the bootstrapped $mean_{std}$ value across all objects. Threshold columns give the fraction of targets with IoU strictly greater than the indicated threshold.}
\label{tab:main_results}
\resizebox{\textwidth}{!}{%
\begin{tabular}{@{}llcccccccc@{}}
\toprule
\multirow{2}{*}{\textbf{Model}} & \multirow{2}{*}{\textbf{Effort}} & \multicolumn{4}{c}{\textbf{All targets}} & \multicolumn{4}{c}{\textbf{Submitted targets}} \\
\cmidrule(lr){3-6}\cmidrule(l){7-10}
& & \textbf{Mean IoU} & \textbf{$>0\%$} & \textbf{$>50\%$} & \textbf{$>90\%$} & \textbf{Mean IoU} & \textbf{$>0\%$} & \textbf{$>50\%$} & \textbf{$>90\%$} \\
\midrule
\includegraphics[height=0.9em]{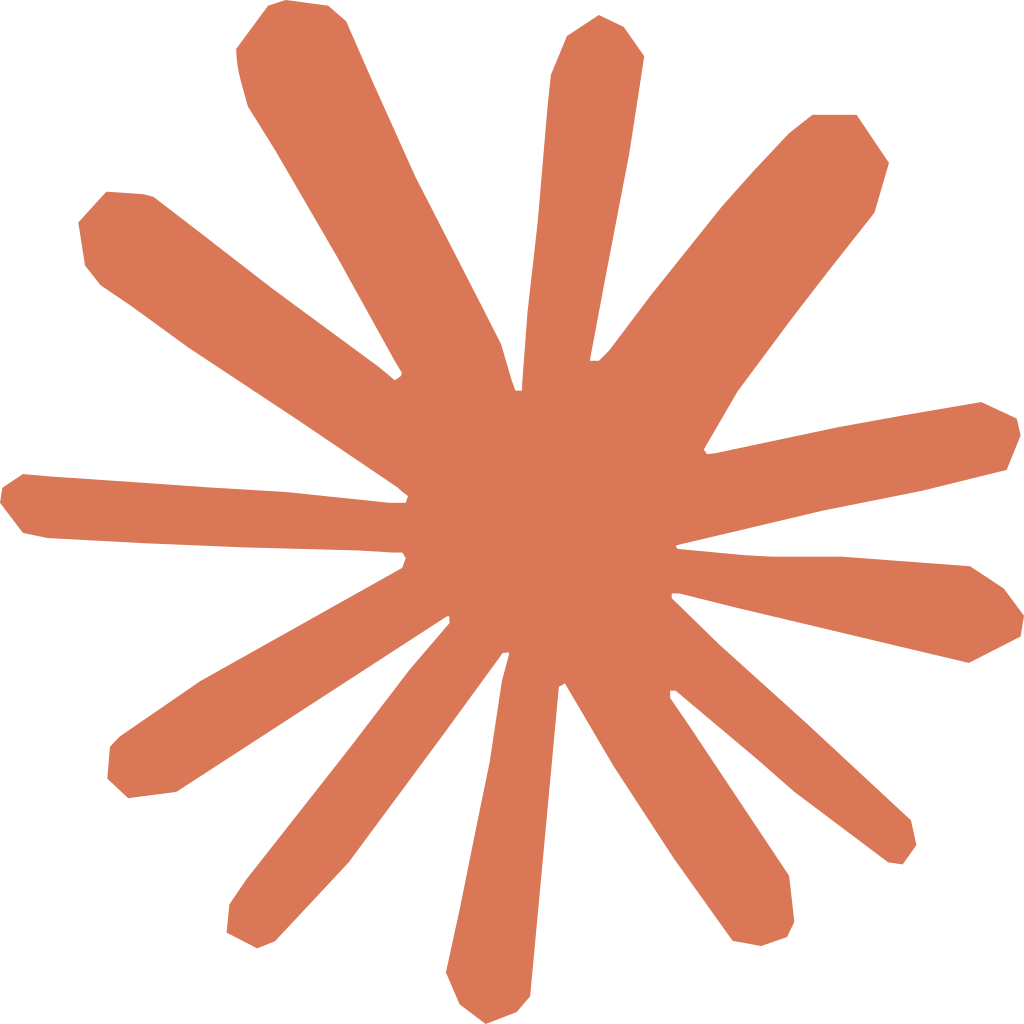} Claude Opus~5.5 & \textit{xhigh} & $22.52_{0.36}$ & $66.18_{1.24}$ & $18.36_{0.64}$ & $0.19_{0.01}$ & $22.58_{0.36}$ & $66.34_{1.22}$ & $18.41_{0.64}$ & $0.19_{0.01}$ \\
\includegraphics[height=0.9em]{figs/claude-color.png} Claude Sonnet~5.5 & \textit{xhigh} & $18.14_{0.26}$ & $62.42_{1.29}$ & $10.61_{0.44}$ & $0.06_{0.00}$ & $18.14_{0.26}$ & $62.42_{1.29}$ & $10.61_{0.44}$ & $0.06_{0.00}$ \\
\includegraphics[height=0.9em]{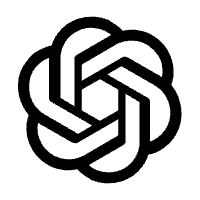} GPT-6 Astra & \textit{xhigh} & $15.50_{0.31}$ & $48.76_{2.09}$ & $10.96_{0.44}$ & $0.13_{0.00}$ & $21.65_{0.37}$ & $68.08_{1.38}$ & $15.30_{0.75}$ & $0.18_{0.01}$ \\
\includegraphics[height=0.9em]{figs/openai-fill.png} GPT-6.1 Sol & \textit{xhigh} & $15.04_{0.28}$ & $50.45_{1.91}$ & $9.47_{0.46}$ & $0.06_{0.00}$ & $19.01_{0.27}$ & $63.77_{1.35}$ & $11.97_{0.59}$ & $0.08_{0.00}$ \\
\includegraphics[height=0.9em]{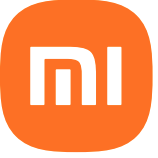} MiMo-V2.6-Pro & \textit{xhigh} & $8.68_{0.28}$ & $37.65_{2.65}$ & $3.76_{0.21}$ & $0.00_{0.00}$ & $8.68_{0.28}$ & $37.65_{2.65}$ & $3.76_{0.21}$ & $0.00_{0.00}$ \\
\includegraphics[height=0.9em]{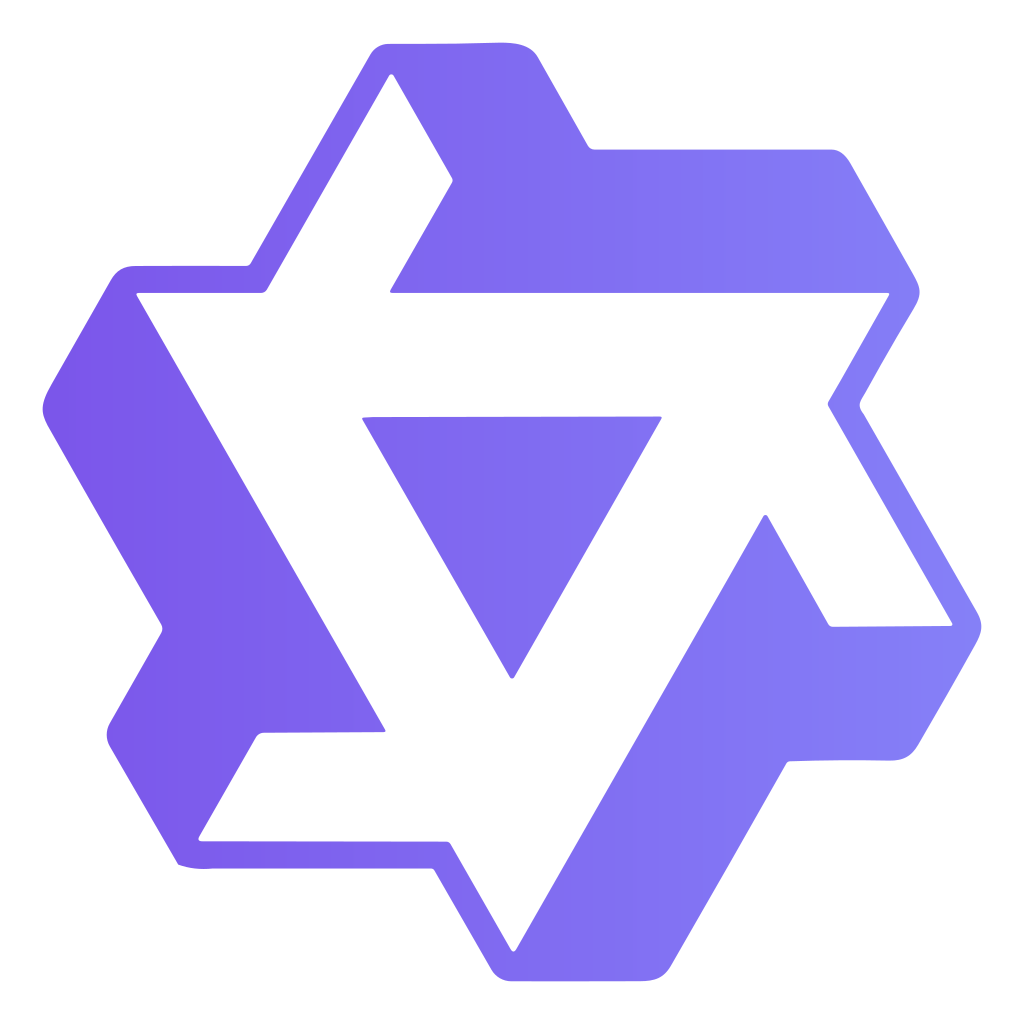} Qwen3.8-Max & \textit{xhigh} & $4.87_{0.19}$ & $22.06_{2.07}$ & $2.46_{0.14}$ & $0.00_{0.00}$ & $4.87_{0.19}$ & $22.06_{2.07}$ & $2.46_{0.14}$ & $0.00_{0.00}$ \\
\includegraphics[height=0.9em]{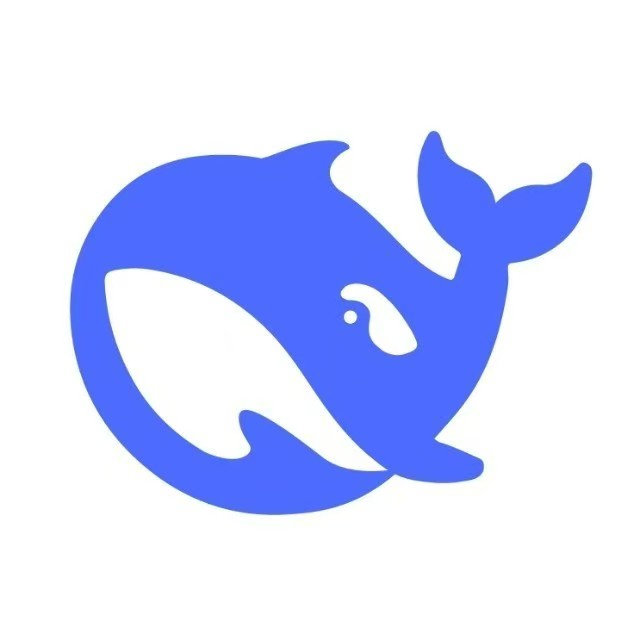} DeepSeek-V4.1-Flash & \textit{xhigh} & $5.49_{0.15}$ & $25.40_{2.00}$ & $2.02_{0.08}$ & $0.00_{0.00}$ & $5.49_{0.15}$ & $25.40_{2.00}$ & $2.02_{0.08}$ & $0.00_{0.00}$ \\
\bottomrule
\end{tabular}
}
\end{table*}

\paragraph{Triangulation.}
For multi-view triangulation~\cite{hartley2004}, the agent can mark corresponding points in at least two acquired images and query \texttt{triangulate\_\allowbreak observations} for a 3D estimate.
The tool minimizes the sum of squared distances to the corresponding viewing lines.
A query requires at least one camera pair separated by 5\,cm or more, a normal-system condition number no greater than $10^6$, and a solution in front of every contributing camera.
The returned point is accompanied by perpendicular ray residuals and the maximum pairwise ray angle.
These diagnostics describe the consistency of the marked image points.
The agent must identify corresponding features and infer how the estimated surface location relates to the object's box center.

\paragraph{Map updates.}
The agent adds or revises centers through \texttt{submit\_positions}.
An accepted submission uses a known target identifier and three finite coordinates; it replaces the previous estimate for that identifier.
Feedback reports acceptance or rejection and the current number of submitted targets.
Ground-truth positions and reference-based scores are not returned through the tools.

The agent autonomously selects its tool calls. Each episode starts with one RGB observation and allows at most 120 model turns. A turn consists of a model response and its requested tool calls.
\subsection{Metrics Based on 3D IoU}
\label{sec:localization_scores}

For each submitted identifier, the agent predicts an object center. We translate
the identifier's reference AABB to this center, preserving its reference side
lengths, and compute its 3D IoU (intersection over union) with reference boxes in the same category.
Predictions and references are matched one-to-one. Mean IoU uses the assignment
that maximizes total IoU. For each threshold
$\tau\in\{0,0.5,0.9\}$, we independently maximize the number of matched pairs
with $\mathrm{IoU}>\tau$, using total IoU to break ties. We report mean IoU and
the rates above these three thresholds; the assignments may differ across
metrics.

% Preserve the intended section number while earlier sections are incomplete.
\setcounter{section}{3}

\begin{figure}
    \centering
    \includegraphics[width=1\linewidth]{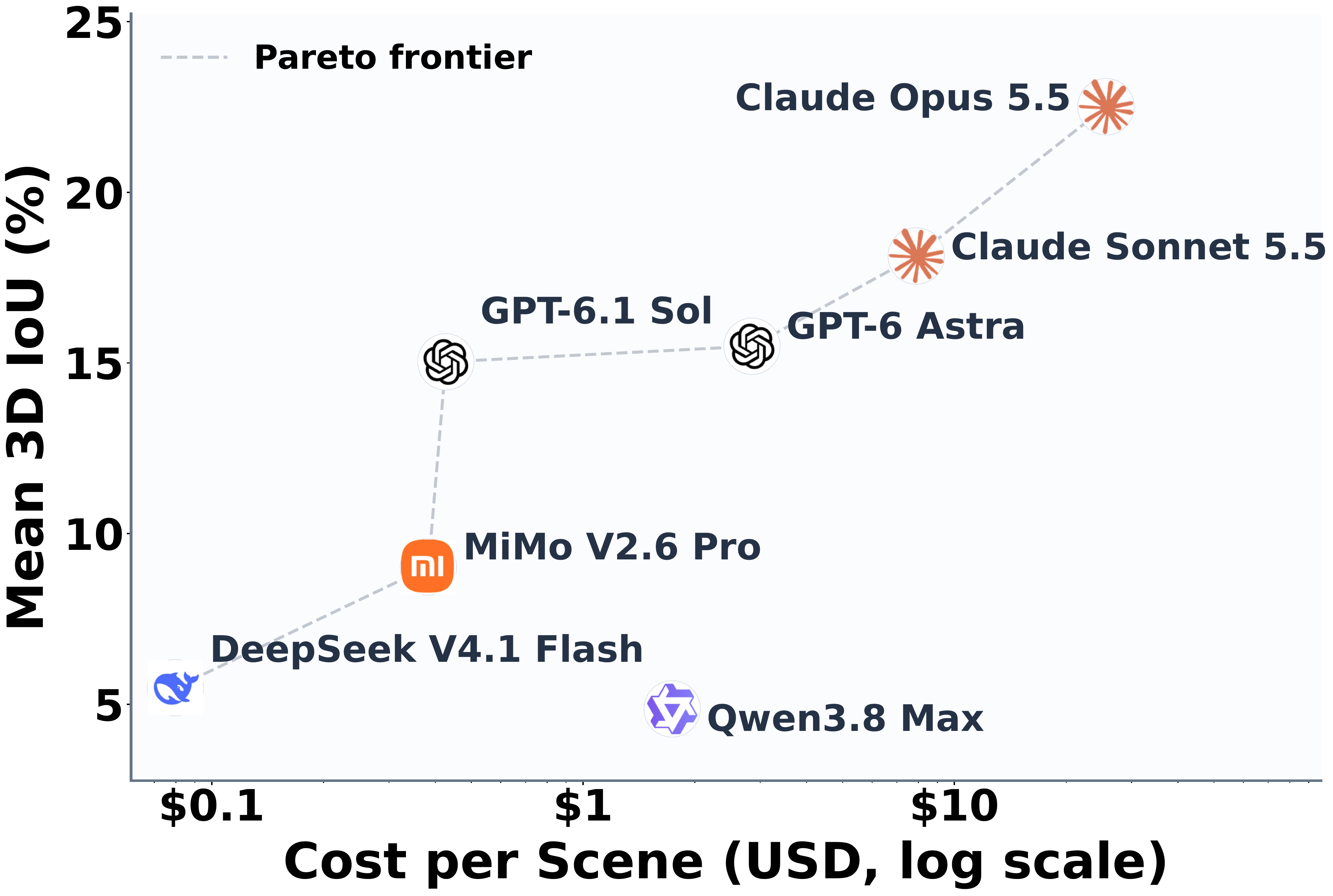}
    \caption{All-target mean IoU versus reported cost per scene, with the pareto frontier.}
    \label{fig:tradeoff}
\end{figure}

\section{Experiments}
\label{sec:experiments}
\begin{figure*}
    \centering
    \includegraphics[width=1\linewidth]{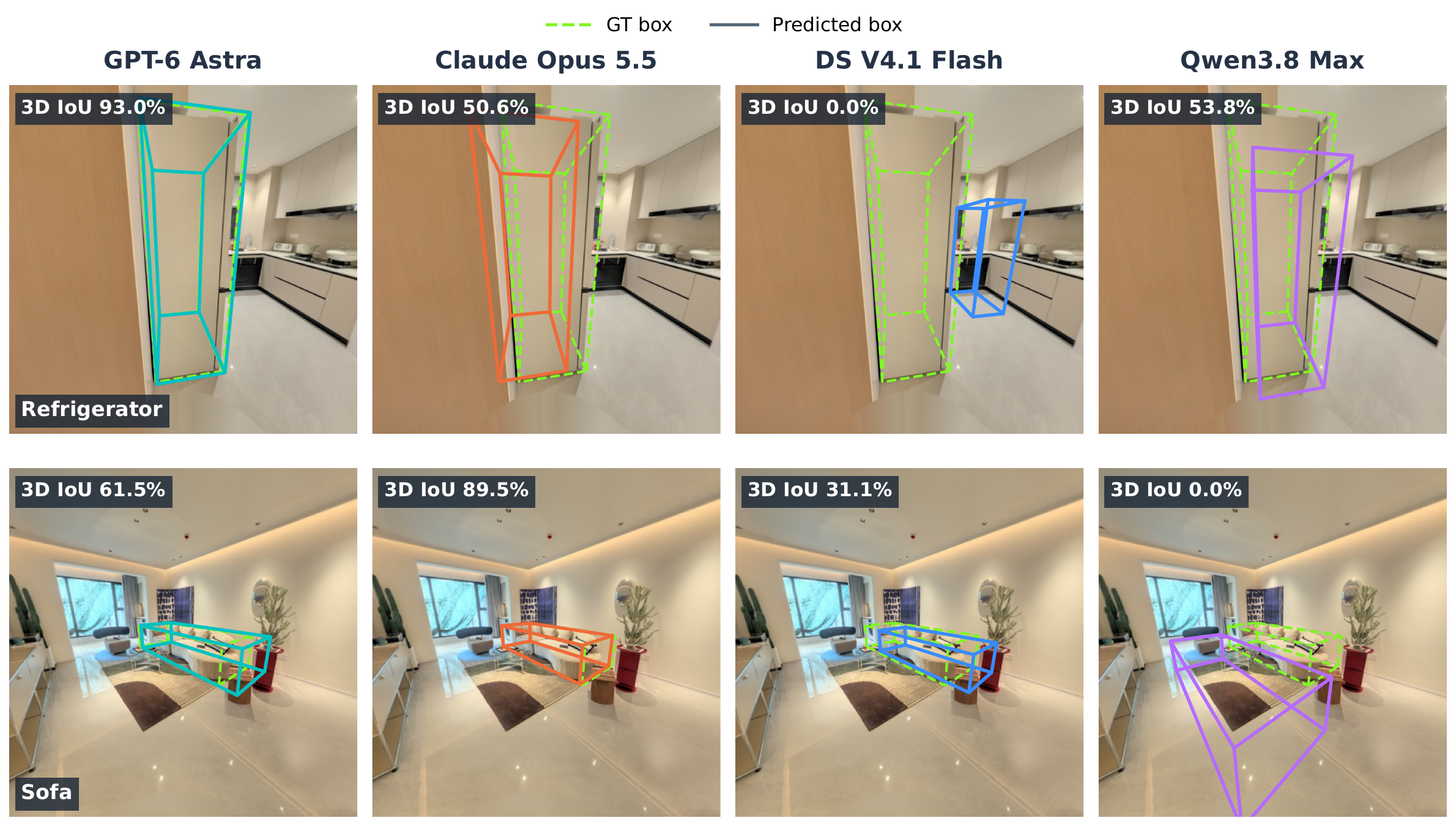}
    \caption{Examples of 3D IoU on SpatialScout-V2. Green boxes show reference objects; the other colors show the corresponding models' predicted boxes.}
    \label{fig:result_visulization}
\end{figure*}

\begin{figure}
    \centering
    \includegraphics[width=1\linewidth]{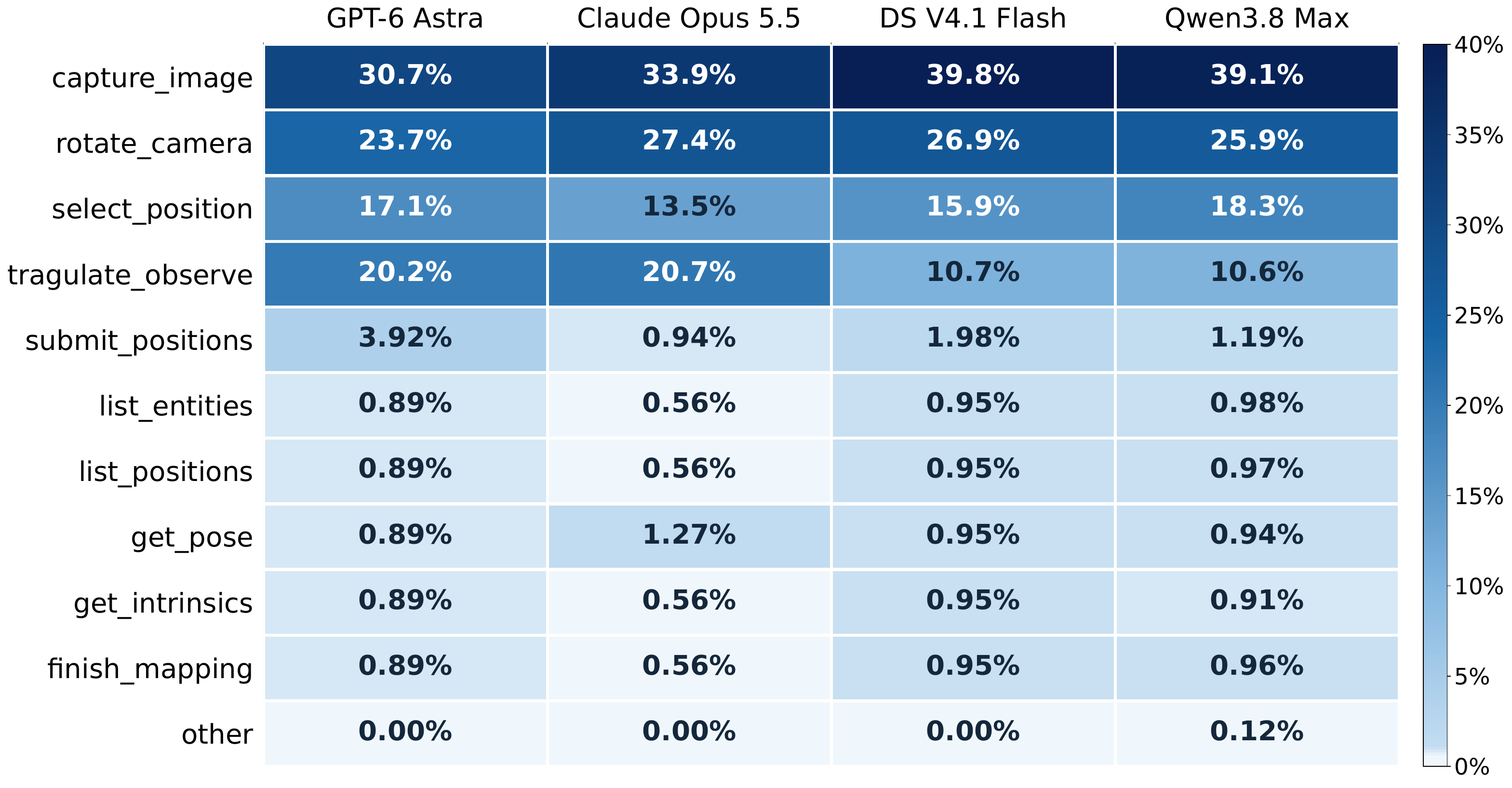}
    \caption{Tool frequency of four evaluated models.}
    \label{fig:tool_freq}
\end{figure}

\subsection{Experimental Setup}
\label{sec:experiments_setup}

We evaluate seven VLM agents on the 100-scene test catalog:
Claude Opus~5.5~\cite{claude2026opus55},
Claude Sonnet~5.5~\cite{claude2026sonnet55},
GPT-6 Astra~\cite{openai2026gpt6astra},
GPT-6.1 Sol~\cite{openai2026gpt61sol},
MiMo-V2.6-Pro~\cite{xiaomi2026mimov26},
Qwen3.8-Max~\cite{qwen2026-38max}, and
DeepSeek-V4.1-Flash~\cite{ds2026-41flash}.
All runs use the \texttt{xhigh} reasoning setting and follow the shared prompt and tool interface in Section~\ref{sec:evaluation_protocol}.
For each scene, agents receive the same target inventory, recorded camera positions, and initial RGB view.
They acquire perspective observations at these positions and submit or revise object centers under the 120-turn limit, with no separate image-count limit.
All evaluated episodes used at most 100 model turns.
We score runs whose final executed tool is a successful \texttt{finish\_mapping} call and whose saved reference and prediction files satisfy the scene manifest.
Partial or empty maps remain eligible and are scored accordingly.
Incomplete attempts are rerun from the beginning, and the runner retains one current completed output per model--scene pair.

Table~\ref{tab:main_results} reports object-micro mean IoU and hit rates at the three strict thresholds defined in Section~\ref{sec:localization_scores}.
All-target scores include omissions, while submitted-target scores describe the estimates in each agent's final map.
We also examine cost per scene, tool-use distributions, repeated operations, and examples of the predicted object locations.

\subsection{Results}
\label{sec:experiments_results}

\paragraph{Overall localization.}
Claude Opus~5.5 achieves the highest all-target mean IoU at 22.52\%, followed by Claude Sonnet~5.5 at 18.14\%.
GPT-6 Astra and GPT-6.1 Sol reach 15.50\% and 15.04\%, respectively.
For Opus, the fraction of targets with positive overlap, 66.18\%, falls to 18.36\% at $\mathrm{IoU}>0.5$ and 0.19\% at $\mathrm{IoU}>0.9$.
The large drop across thresholds shows that placing an object in roughly the correct region remains much easier than aligning its center precisely relative to its spatial extent.
Fine localization remains difficult among submitted targets as well: no model exceeds a hit rate of 0.19\% at $\mathrm{IoU}>0.9$.

\paragraph{Cost and accuracy.}
Fig.~\ref{fig:tradeoff} compares localization performance with the reported cost per scene.
Claude Opus~5.5 attains the highest mean IoU and incurs the highest cost, while DeepSeek-V4.1-Flash occupies the lowest-cost end of the comparison.
GPT-6.1 Sol achieves a mean IoU 0.46 percentage points below Astra at substantially lower cost.

\paragraph{Qualitative examples.}
Fig.~\ref{fig:result_visulization} illustrates the localization differences measured by 3D IoU.
For the refrigerator, Astra achieves 93.0\% IoU, while Opus reaches 50.6\% and DeepSeek's displaced estimate has no overlap with the reference.
For the sofa, Claude-Opus-5.5 achieves 89.5\% IoU, compared with 61.5\% for GPT-6-Astra, 31.1\% for DeepSeek-V4.1 Flash, and zero for Qwen3.8-max.

\begin{figure*}
    \centering
    \includegraphics[width=1\linewidth]{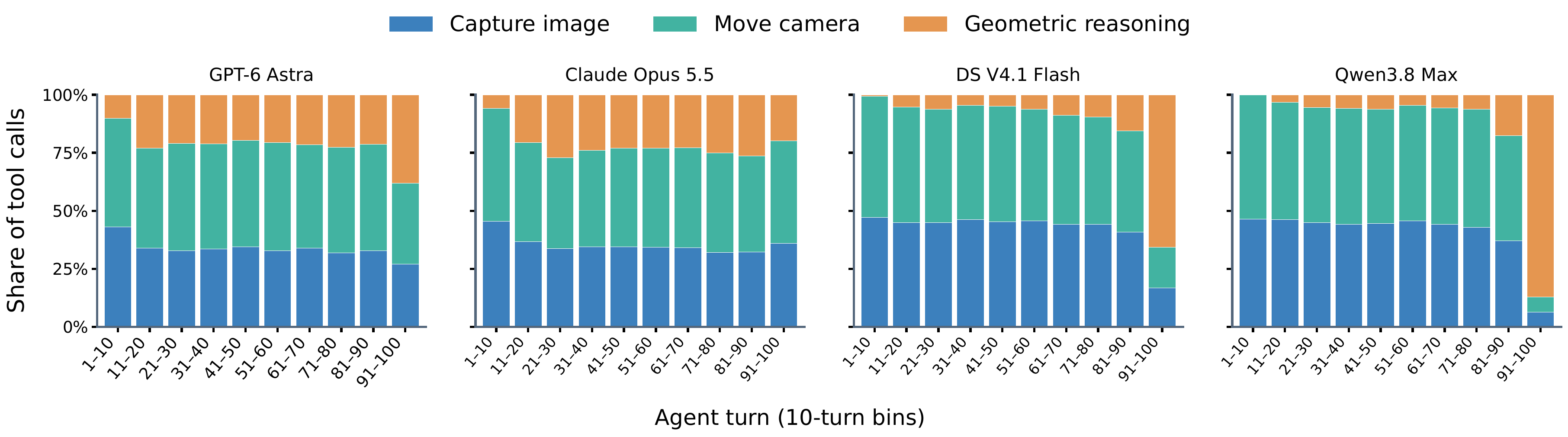}
    \caption{Tool-call distributions in ten-turn intervals for GPT-6 Astra, Claude Opus~5.5, DeepSeek-V4.1-Flash, and Qwen3.8-Max. Calls are grouped into \textit{Capture Image, Move Camera, and Geometric Reasoning}.}
    \label{fig:tool-analysis}
    \vspace{-2mm}
\end{figure*}
\subsection{Analysis}

\paragraph{Localization and coverage.}
GPT-6 Astra reaches a submitted-target mean IoU of 21.65\%, close to Opus at 22.58\%, but their all-target scores differ by 7.02 percentage points.
Astra's mean IoU falls by 6.15 points when omissions are included, compared with only 0.06 points for Opus.
This difference reflects the effect of inventory coverage on the aggregate score.
GPT-6.1 Sol shows a similar reduction, from 19.01\% over submitted targets to 15.04\% over all targets.
Conversely, MiMo, Qwen, and DeepSeek have identical all-target and submitted-target scores at the reported precision, yet their mean IoUs are only 8.68\%, 4.87\%, and 5.49\%.
Their low overlap persists despite broad submission, showing why coverage and localization accuracy must be considered together.

\paragraph{Tool-use patterns.}
Fig.~\ref{fig:tool_freq} shows how the four agents distribute tool calls across observation, camera control, and mapping. Triangulation accounts for 20.2\% of calls for Astra and 20.7\% for Opus, compared with 10.7\% for DeepSeek and 10.6\% for Qwen. By contrast, DeepSeek and Qwen devote larger shares of their calls to image capture (39.8\% and 39.1\%, respectively) than Astra and Opus (30.7\% and 33.9\%). These aggregate patterns suggest that the higher-scoring models place greater relative emphasis on geometric estimation across views.

Fig.~\ref{fig:tool-analysis} shows how tool use varies across ten-turn intervals. During the first 80 turns, DeepSeek and Qwen allocate most calls to image capture and camera control, with a smaller share devoted to geometric reasoning. Their geometric-reasoning share rises in the final two intervals. Astra and Opus maintain a higher geometric-reasoning share through most of the episode. These distributions show how the four models divide their interactions between acquiring observations and estimating object locations.
\begin{figure}
    \centering
    \includegraphics[width=1\linewidth]{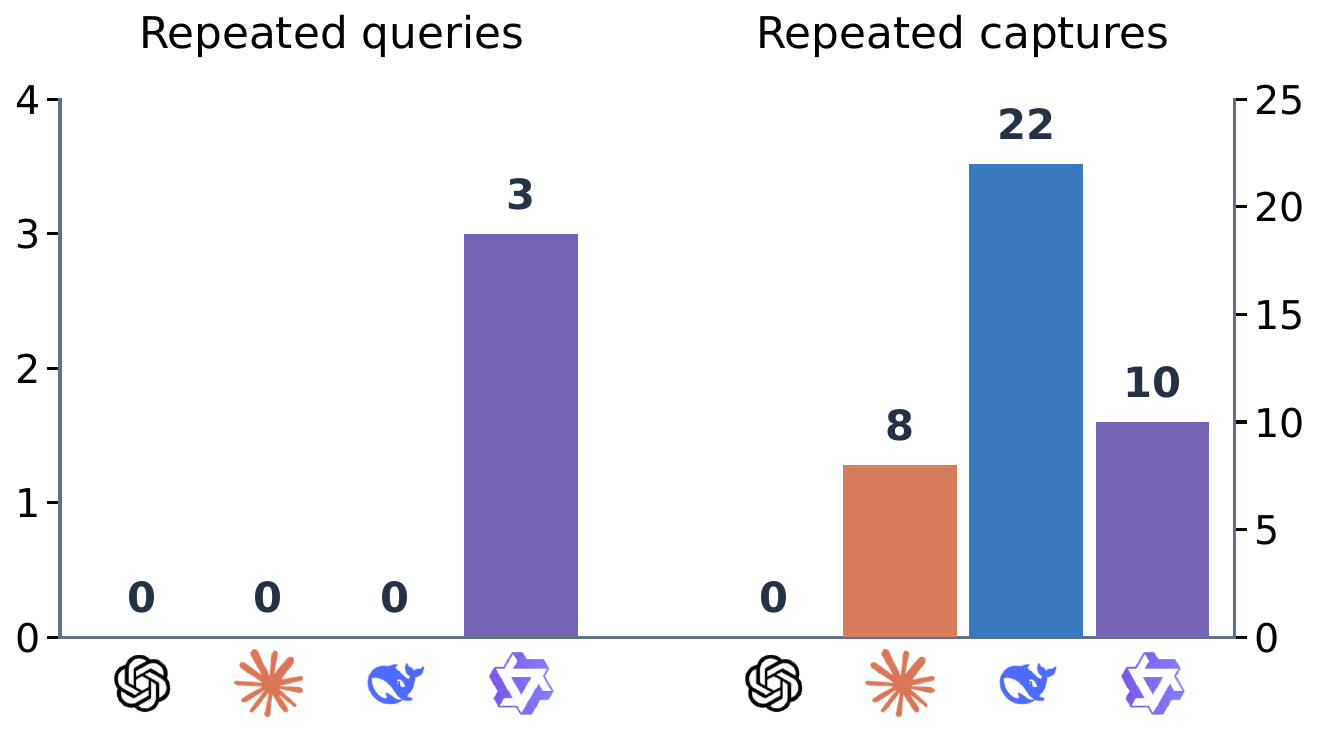}
    \caption{Repeated queries (left) and repeated captures (right) for the four analyzed models.}
    \label{fig:redundancy}
\end{figure}

\paragraph{Repeated operations.}
Fig.~\ref{fig:redundancy} reports repeated operations during inference. Repeated queries retrieve unchanged camera information, while repeated captures reproduce an identical image. Astra and Opus show fewer repeated captures than DeepSeek and Qwen. These operations add tool outputs to the context without providing additional scene evidence. Repeated operations only consume context without providing any new information.

\begin{table}[t]
\centering
\small
\caption{Success factors and failure patterns across 70 trajectories (6 successful and 64 unsuccessful). Success requires all-target mean IoU $>50\%$.}
\label{tab:trajectory_causes}
\resizebox{\columnwidth}{!}{%
\begin{tabular}{@{}lp{0.62\columnwidth}r@{}}
\toprule
\textbf{Outcome} & \textbf{Factor} & \textbf{Total} \\
\midrule
Success & \textbf{S}ufficient multi-view grounding & 6 \\
 & \textbf{I}ncremental submission with continued coverage expansion & 0 \\
 & \textbf{C}ontinued refinement & 4 \\
\midrule
Failure & \textbf{I}ncomplete object coverage & 6 \\
 & \textbf{S}uspected correspondence errors despite low triangulation residuals & 14 \\
 & \textbf{U}nresolved inconsistencies in geometric measurements & 21 \\
 & \textbf{E}arly termination before coverage & 6 \\
 & \textbf{P}osition estimates lacking triangulation support & 39 \\
 & \textbf{E}xtensive observation with limited localization progress & 14 \\
 & \textbf{L}ate-stage refinement without recovering missing objects & 1 \\
\bottomrule
\end{tabular}}
\end{table}
\paragraph{Failure Analysis}
We select 10 scenes and mannually analyze their trajectories from the 7 models in Table~\ref{tab:main_results}. Table~\ref{tab:trajectory_causes} summarizes 70 analyzed trajectories, of which six achieve an all-target mean IoU above 50\%. All six successful trajectories use observations from multiple views to support localization, and four continue to refine their estimates. None combines incremental submission with continued expansion of object coverage. Among the 64 unsuccessful trajectories, position estimates without triangulation support are the most frequent pattern (39/64), followed by unresolved geometric inconsistencies (21/64). Suspected correspondence errors despite low triangulation residuals and extensive observation with limited localization progress each appear in 14 trajectories. Incomplete coverage and early termination each occur in six trajectories, while one focuses on late-stage refinement without recovering missing objects. The most frequent failure patterns concern converting observations into geometrically supported object locations.

\section{Related Work}
\label{sec:related_work}

\paragraph{Spatial reasoning and embodied evaluation.}
VSR tests whether image--caption pairs express correct spatial relations~\cite{vsr}; BLINK evaluates relative depth, visual correspondence, and multi-view reasoning from supplied images~\cite{blink}.
VSI-Bench evaluates spatial understanding from supplied videos~\cite{vsi}, while Metric-Bench tests absolute metric inference using objects with known reference measurements~\cite{metricbench}.
ScanQA evaluates question answering from supplied 3D scans~\cite{scanqa}; SQA3D requires reasoning from an agent's text-described position and orientation~\cite{sqa3d}.
OpenEQA supports both episodic-memory and active-exploration question answering~\cite{openeqa}, ESI-Bench evaluates spatial questions requiring perception and action~\cite{esi}, and S$^3$-Bench combines streaming observations, memory, and further exploration~\cite{s3}.
E3VS-Bench studies viewpoint-dependent visual search in reconstructed real-world scenes~\cite{e3vs}.
Theory of Space directly probes the construction and revision of cognitive maps in procedurally generated grid environments~\cite{theoryofspace}.
SpatialScout-V2 evaluates the accuracy and completeness of object-center maps built through autonomous evidence acquisition.

\paragraph{Object-level grounding and geometric mapping.}
ScanRefer localizes a language-referred object in a supplied 3D scan~\cite{scanrefer}, whereas ConceptFusion and ConceptGraphs fuse observations across views into semantically enriched three-dimensional representations~\cite{conceptfusion,conceptgraphs}.
AnchorVLN combines vision-language reasoning with LiDAR-derived geometry for object reference and navigation~\cite{anchorvln}.
Our agents receive calibrated RGB views and camera poses, with reference depth and object boxes withheld from the interaction interface.
They must associate physical instances across observations and distinguish a triangulated visible feature from the center of an object's spatial extent.
DUSt3R and VGGT provide complementary approaches to recovering geometric structure from images~\cite{dust3r,vggt}.
With camera geometry supplied, SpatialScout-V2 evaluates how agents select and integrate observations to estimate object centers.

\paragraph{Active perception and reconstruction.}
For embodied question answering, Explore until Confident guides exploration with semantic maps and calibrates answer confidence to determine when to stop~\cite{exploreeqa}.
3D-Mem constructs memory snapshots from posed RGB-D observations and uses frontier snapshots for exploration~\cite{mem3d}.
GenNBV learns a generalizable next-best-view policy~\cite{gennbv}, while NARUTO couples exploration with uncertainty-aware neural reconstruction~\cite{naruto}.
AREA3D combines feed-forward geometric perception and vision-language guidance for active reconstruction~\cite{area3d}.
FisherRF uses Fisher information to score candidate views for radiance-field reconstruction~\cite{fisherrf}.
Agentic inverse-graphics benchmarks, including IR3D-Bench and SceneActBench, also assess explicit geometric outputs through scene-parameter estimation or construction of object geometry~\cite{ir3d,sceneact}.
SpatialScout-V2 evaluates object-center accuracy and inventory coverage with shared camera and triangulation tools.
Recorded imagery provides fixed visual content while allowing agents to adaptively select observations and revise positions.

\paragraph{Real-world scene resources and SpatialScout.}
ScanNet and Matterport3D established real indoor RGB-D scans, registered observations, and semantic annotations as foundations for three-dimensional perception~\cite{scannet,matterport3d}.
The Active Vision Dataset benchmarks object detection and active recognition using densely sampled RGB-D views of real indoor scenes~\cite{ammirato2017avd}.
The Matterport3D Simulator and Room-to-Room dataset support instruction-following navigation through connected panoramic viewpoints captured in real buildings~\cite{anderson2018vln}.
Realsee3D provides real and synthetic multi-view indoor panoramas~\cite{Li2025realsee3d_data}; Argus uses this resource to study metric panoramic reconstruction of camera poses, depth, and point clouds~\cite{li2026argus}.
SpatialScout introduced the active object-center mapping task in simulator-rendered indoor scenes~\cite{spatialscout}.
SpatialScout-V2 retains SpatialScout's inventory-based mapping task and omission-aware evaluation, using real residential photographs and human-reviewed reference instances derived from registered RGB-D captures.
Agents select recorded scan positions while non-target objects remain visible with their original occlusions.

\section{Conclusion}
We introduce SpatialScout-V2, a benchmark for evaluating active view selection and 3D object localization from recorded images of real homes. The reference annotations are manually verified. During evaluation, VLM agents select recorded viewpoints, adjust viewing directions, acquire observations, and estimate target object centers in a common metric frame.

Our results show that broad target submission can coexist with low localization accuracy, while target omissions further reduce all-target scores. Tool-use analysis reveals differences in how agents allocate interactions to image capture and geometric estimation, together with repeated operations and early termination. SpatialScout-V2 provides a basis for developing and evaluating models and agent harnesses that integrate visual evidence into accurate and complete object-center maps.

\section*{LLM Usage Statement}

We used GPT (via Codex) to assist with code development, manuscript drafting and revision, and reference checking, and Gemini to assist with grammar checking. The authors manually reviewed the code and manuscript and take full responsibility for all submitted content (paper, code, and other materials).
{
    \small
    \bibliographystyle{ieeenat_fullname}
    \bibliography{main}
}


\clearpage
\appendix
\section{Supplementary Details}
\label{app:protocol_details}

This section specifies the target manifests, information available to the agent, trajectory accounting, and aggregation of the eight localization scores.
The prompts and tool instructions are reproduced in Appendix~\ref{app:prompts}.

\subsection{Scene Capture and Target Manifests}
\paragraph{Registered real captures.}
Realsee's capture system combines RGB cameras and LiDAR with nearly coincident projection centers~\cite{li2026argus}.
Images acquired at multiple orientations are stitched in the RAW domain, and multiple exposures are combined into HDR panoramas.
The scan poses are refined through global registration using image features and 3D observations, with manual correction where needed.
The released validity masks identify blind regions near the top and bottom of the panoramas.
These registered captures provide the common metric frame used for multi-view annotation and evaluation.

\paragraph{Fixed target inventory.}
Each scene's selected objects are stored in an \texttt{objects\_3dbox.json} manifest.
The evaluator uses every entry in this file, without further sampling during a model run.
The same manifest defines the inventory for all models, while objects outside the inventory remain visible in the photographs.
An entity identifier combines the object's category name and its index in the manifest; its matching group is the exact category name.
The localization target is the midpoint of the box's spatial extent, as defined in Eq.~\ref{eq:target_center}.
A point on the visible surface or at the center of a 2D silhouette can occupy a different 3D location.

\paragraph{Information available to the agent.}
The public inventory contains each target's identifier, category, and matching group.
The agent can also query the available camera-position identifiers and coordinates, the current camera pose, and the intrinsic matrix.
Each captured frame has a saved acquisition pose and calibration, which the triangulation tool uses even after the camera has moved.
Depth maps, segmentation masks, reference boxes, annotation confidences, visibility lists, and floor labels are used for benchmark preparation and evaluation and are not included in the agent's observations or tool responses.

\subsection{Trajectory Execution and Completion}
\label{sec:run_accounting}
A model response can request multiple tools, which execute sequentially before their outputs are supplied to the next model turn.
The turn budget is specified in Sec.~\ref{sec:interaction}; there is no separate cap on the number of captured images.
The conversation retains previous observations and tool outputs.
Transient API failures are retried within the current turn.
The evaluator sends reminders when ten and three model turns remain.
After either of the first two consecutive responses without a tool call, it asks the agent to continue; a third consecutive response without a tool call ends the attempt.
Recoverable tool errors are returned to the agent so that it can revise its request.

The scoring script accepts a trajectory when its final executed tool is \texttt{finish\_mapping} with \texttt{finished=true}, its reference and prediction files are present, and the reference count agrees with the saved target count.
A completed map can be partial or empty.
Its unlocalized targets remain in the denominator of the all-target scores.
Attempts that end without satisfying these completion requirements are excluded from aggregation.
All eight metrics for a model use the same set of eligible scenes.

Before reusing a completed scene, the batch runner checks the saved scene identifiers and target counts against the input manifest and verifies that the prediction and metric files belong to the completed attempt.
Incomplete scenes are rerun from the beginning.
The runner retains one current output per model--scene pair, including reference boxes, final predictions, and the tool trace needed to recompute the scores.

\subsection{Score Definitions and Aggregation}
\label{sec:metric_aggregation}
\paragraph{Pairwise overlap.}
For submitted identifier $j$, let $B_j$ be its reference AABB, $\mathbf{p}_j$ its reference center, and $\widehat{\mathbf{p}}_j$ its submitted center.
The evaluator translates that box to the submitted center and computes its overlap with a candidate reference $B_i$ in the same category:
\begin{equation}
    \begin{aligned}
        \widehat{B}_j &= B_j+(\widehat{\mathbf{p}}_j-\mathbf{p}_j),\\
        J_{ji} &= \frac{\operatorname{Vol}(\widehat{B}_j\cap B_i)}{\operatorname{Vol}(\widehat{B}_j\cup B_i)}.
    \end{aligned}
    \label{eq:translated_box_iou}
\end{equation}
The translated box retains the reference side lengths associated with identifier $j$ throughout matching.
A pair with zero union volume receives zero IoU.
The one-to-one assignments for mean IoU and each strict threshold follow Sec.~\ref{sec:localization_scores}.
The evaluator solves these rectangular assignment problems with SciPy's \texttt{linear\_sum\_assignment} routine, which uses a modified Jonker--Volgenant algorithm~\cite{crouse2016}.

\paragraph{All-target and submitted-target scores.}
For eligible scene $s$, let $N_s$ be the number of inventory targets and $M_s$ the number of distinct submitted identifiers in the final map.
Let $Q_s$ be the sum of matched IoUs under the maximum-total-IoU assignment, and let $H_s(\tau)$ be the number of hits under the assignment for threshold $\tau$.
The object-micro scores reported in Table~\ref{tab:main_results} pool these quantities across scenes:
\begin{equation}
    \overline{J}_{\mathrm{all}}=\frac{\sum_s Q_s}{\sum_s N_s},
    \qquad
    \overline{J}_{\mathrm{sub}}=\frac{\sum_s Q_s}{\sum_s M_s},
    \label{eq:mean_iou}
\end{equation}
\begin{equation}
    R_{\mathrm{all}}(\tau)=\frac{\sum_s H_s(\tau)}{\sum_s N_s},
    \quad
    R_{\mathrm{sub}}(\tau)=\frac{\sum_s H_s(\tau)}{\sum_s M_s}.
    \label{eq:overlap_rates}
\end{equation}
We use $\tau\in\{0,0.5,0.9\}$ and report all eight scores as percentages.
The comparisons are strict: a pair with IoU exactly equal to the threshold is not a hit.

All-target scores give equal weight to each inventory object and zero credit to unmatched references.
Submitted-target scores give equal weight to each accepted estimate in the final map; this submitted subset is chosen by the agent and can differ across models.
An empty completed map contributes zero IoU and zero hits while retaining its full inventory in the all-target denominator.
Ratios with a zero denominator are set to zero, and no aggregate report is produced when there are no eligible scenes.

\paragraph{Submission coverage.}
Coverage is the fraction of targets with a submitted center, $C=\sum_s M_s/\sum_s N_s$.
For object-micro aggregation, the shared numerators give
\begin{equation}
    \overline{J}_{\mathrm{all}}=C\overline{J}_{\mathrm{sub}},
    \qquad
    R_{\mathrm{all}}(\tau)=C R_{\mathrm{sub}}(\tau).
    \label{eq:coverage_relation}
\end{equation}
Thus, all-target scores combine inventory coverage with the localization accuracy of submitted estimates.

\section{Agent Prompts and Tool Interface}
\label{app:prompts}

The real-scene evaluator supplies a common instruction and a scene-specific mapping task, followed by the initial RGB image.
The instruction is sent as a developer message through the Responses interface and as the system instruction through the Anthropic and Gemini interfaces.
The text below reproduces the prompts used by the evaluator; angle-bracketed fields are replaced with values from the current scene or trajectory.

\subsection{Initial Instructions}
\paragraph{Agent instruction.}
\begin{quote}
\small\ttfamily\raggedright
\detokenize{You are a visual mapping agent. Use the provided tools and return positions in the dataset world coordinates, in meters. Triangulate using multiple views.}
\end{quote}
\paragraph{Scene task.}
\texttt{<TARGET\_COUNT>} is the number of selected objects.
\texttt{<TARGET\_INVENTORY\_JSON>} is the serialized list of records containing \texttt{id}, \texttt{category}, and \texttt{group}.
\begin{quote}
\small\ttfamily\raggedright
\detokenize{Map only these <TARGET_COUNT> selected target entities, and no other objects: <TARGET_INVENTORY_JSON>. Call list_entities and list_camera_positions. Select only listed camera points with select_camera_position; rotate_camera changes the view. All positions, poses and predictions use dataset world coordinates in meters, z up. Use capture_image and triangulate_observations from multiple points to estimate object centers. Equivalent IDs in the same group are interchangeable. Submit visually supported positions using submit_positions and call finish_mapping.}
\end{quote}
\subsection{Image and Progress Messages}
\paragraph{Image message.}
Each captured image is accompanied by the following text, with \texttt{<FRAME\_NAME>} replaced by its filename, such as \texttt{frame\_0001.png}.
\begin{quote}
\small\ttfamily\raggedright
\detokenize{Camera image <FRAME_NAME>, recorded when capture_image was called. Later camera moves may have changed the current view. Continue exploring or answer the task.}
\end{quote}
\paragraph{Continuation message.}
After either of the first two consecutive responses without a tool call, the evaluator appends this message with the current submission and target counts.
\begin{quote}
\small\ttfamily\raggedright
\detokenize{The mapping task is still active. You have submitted <SUBMITTED_COUNT> of <TARGET_COUNT> target positions. Your last response did not call a tool. Continue with a tool call; if you are done exploring, submit any well-supported positions and call finish_mapping.}
\end{quote}
\paragraph{Budget reminders.}
The evaluator appends the following reminders after tool execution when ten and three model turns remain, respectively.
\begin{quote}
\small\ttfamily\raggedright
\detokenize{Ten model calls remain. Submit all well-supported positions now, then use the remaining calls for corrections. Missing targets count as missing predictions.}
\end{quote}
\begin{quote}
\small\ttfamily\raggedright
\detokenize{Three model calls remain. Submit any final verified positions and call finish_mapping.}
\end{quote}
\subsection{Tool Instructions}
The following descriptions are supplied alongside the prompts as function definitions.
The real-scene interface exposes ten tools.
\paragraph{\texttt{get\_camera\_pose}.}
\begin{quote}
\small\ttfamily\raggedright
\detokenize{Read the camera pose in dataset world coordinates.}
\end{quote}
This tool takes no arguments.
\paragraph{\texttt{get\_camera\_intrinsics}.}
\begin{quote}
\small\ttfamily\raggedright
\detokenize{Read the actual 3x3 camera intrinsic matrix, image size, and clipping planes.}
\end{quote}
This tool takes no arguments.
\paragraph{\texttt{rotate\_camera}.}
\begin{quote}
\small\ttfamily\raggedright
\detokenize{Rotate the camera by relative yaw and pitch in degrees. Positive yaw turns counterclockwise viewed from above; positive pitch looks upward.}
\end{quote}
The arguments \texttt{yaw\_deg} and \texttt{pitch\_deg} are numeric rotation increments in degrees.
\paragraph{\texttt{capture\_image}.}
\begin{quote}
\small\ttfamily\raggedright
\detokenize{Capture the camera's current RGB image. The PNG is saved locally and the image is sent to your vision input.}
\end{quote}
This tool takes no arguments.
\paragraph{\texttt{list\_entities}.}
\begin{quote}
\small\ttfamily\raggedright
\detokenize{List every target entity ID and category in this room. No locations are revealed.}
\end{quote}
This tool takes no arguments.
\paragraph{\texttt{triangulate\_observations}.}
\begin{quote}
\small\ttfamily\raggedright
\detokenize{Estimate one entity's absolute 3D center from the same image point marked in two or more captured frames taken from different camera positions. Returns the least-squares ray intersection and ray mismatch; no ground truth is used.}
\end{quote}
The arguments are an entity \texttt{id} and an \texttt{observations} list with at least two entries. Each entry contains a captured \texttt{frame} filename and a two-number \texttt{pixel\_xy} array.
\paragraph{\texttt{submit\_positions}.}
\begin{quote}
\small\ttfamily\raggedright
\detokenize{Submit object centers in dataset world coordinates in meters. Use exact entity IDs; batches can be updated.}
\end{quote}
The \texttt{predictions} argument is a list of records, each containing an entity \texttt{id} and a three-number \texttt{xyz\_m} array.
\paragraph{\texttt{finish\_mapping}.}
\begin{quote}
\small\ttfamily\raggedright
\detokenize{Finish exploration and score all submitted positions. Missing IDs count as missing predictions.}
\end{quote}
This tool takes no arguments.
\paragraph{\texttt{list\_camera\_positions}.}
\begin{quote}
\small\ttfamily\raggedright
\detokenize{List every available camera point ID and its absolute xyz coordinates in meters.}
\end{quote}
This tool takes no arguments.
\paragraph{\texttt{select\_camera\_position}.}
\begin{quote}
\small\ttfamily\raggedright
\detokenize{Move to an available camera point by its exact ID. Orientation remains unchanged relative to the panorama.}
\end{quote}
The \texttt{id} argument is a camera-position identifier returned by \texttt{list\_camera\_positions}.

\section{More Examples}
\begin{figure*}
    \centering
    \includegraphics[width=1\linewidth]{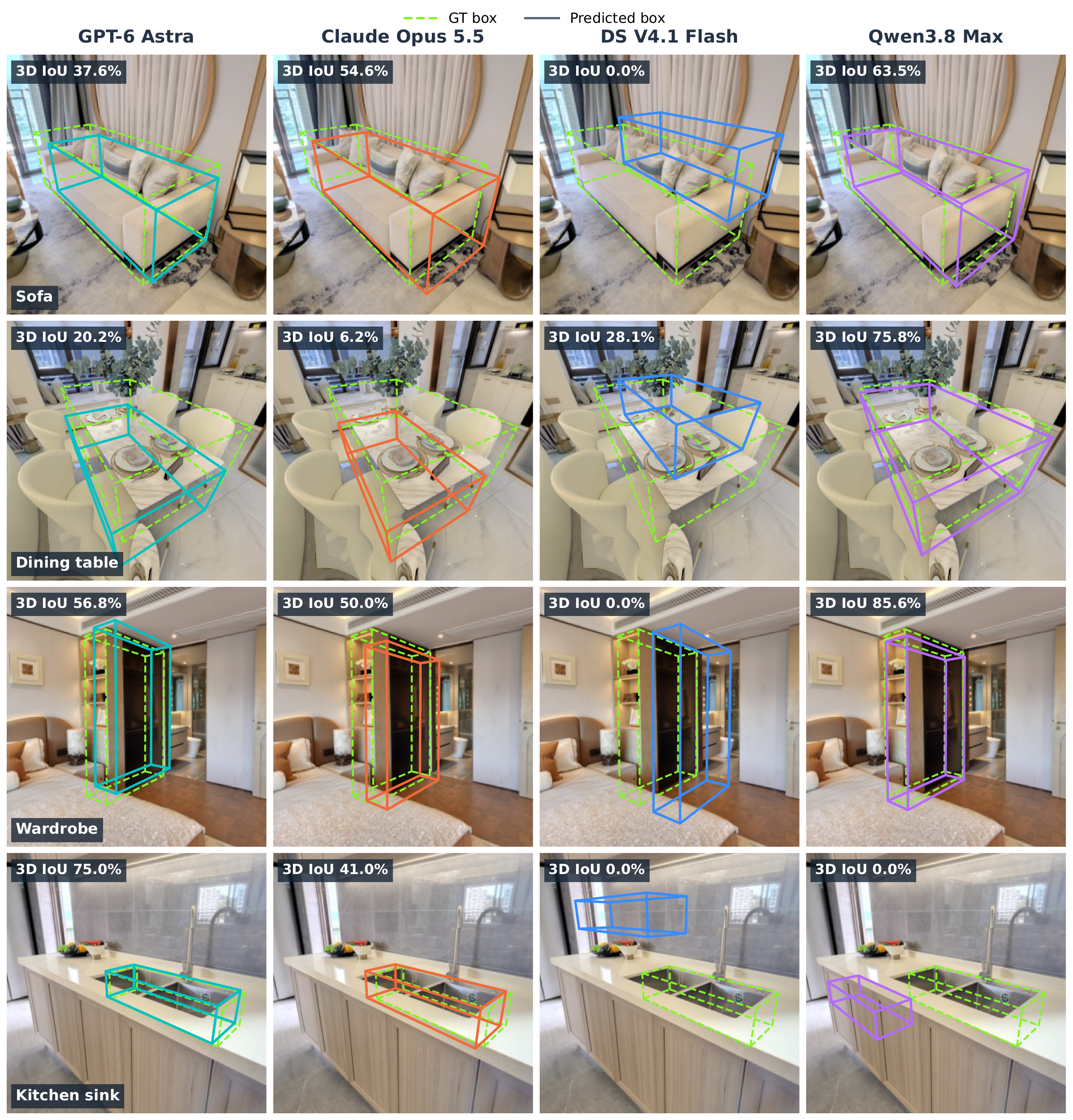}
    \caption{More examples of accurate and inaccurate 3D mapping results. The green box stands for the groundtruth bounding box, while the other colors represent the predicted bounding box by the corresponding model.}
    \label{fig:full-examples}
\end{figure*}

We present 4 extra cases in Fig.~\ref{fig:full-examples}, including 4 objects, sofa, dining table, wardrobe, and kitchen sink.

\end{document}